%% file: main.tex
\documentclass[10pt,twocolumn,letterpaper]{article}

\usepackage[pagenumbers]{cvpr} 

\input{preamble}

\definecolor{cvprblue}{rgb}{0.21,0.49,0.74}
\usepackage[pagebackref,breaklinks,colorlinks,allcolors=cvprblue]{hyperref}

\usepackage{tablefootnote}

\title{SlotVTG: Object-Centric Adapter for Generalizable Video Temporal Grounding}

\author{
Jiwook Han${}^{1}{}^{*}$ \quad
Geo Ahn${}^{1}{}^{*}$ \quad
Youngrae Kim${}^{2}{}^{*}$ \quad
Jinwoo Choi${}^{1}{}^{\dagger}$ \quad \\
${}^{1}$Kyung Hee University\quad ${}^{2}$University of Southern California \\
{\tt\small \{mreraser,ahngeo11,jinwoochoi\}@khu.ac.kr},\;
{\tt\small youngrae@usc.edu}
}

\begin{document}

\twocolumn[{
  \maketitle
  \begin{center}
    \includegraphics[width=\linewidth]{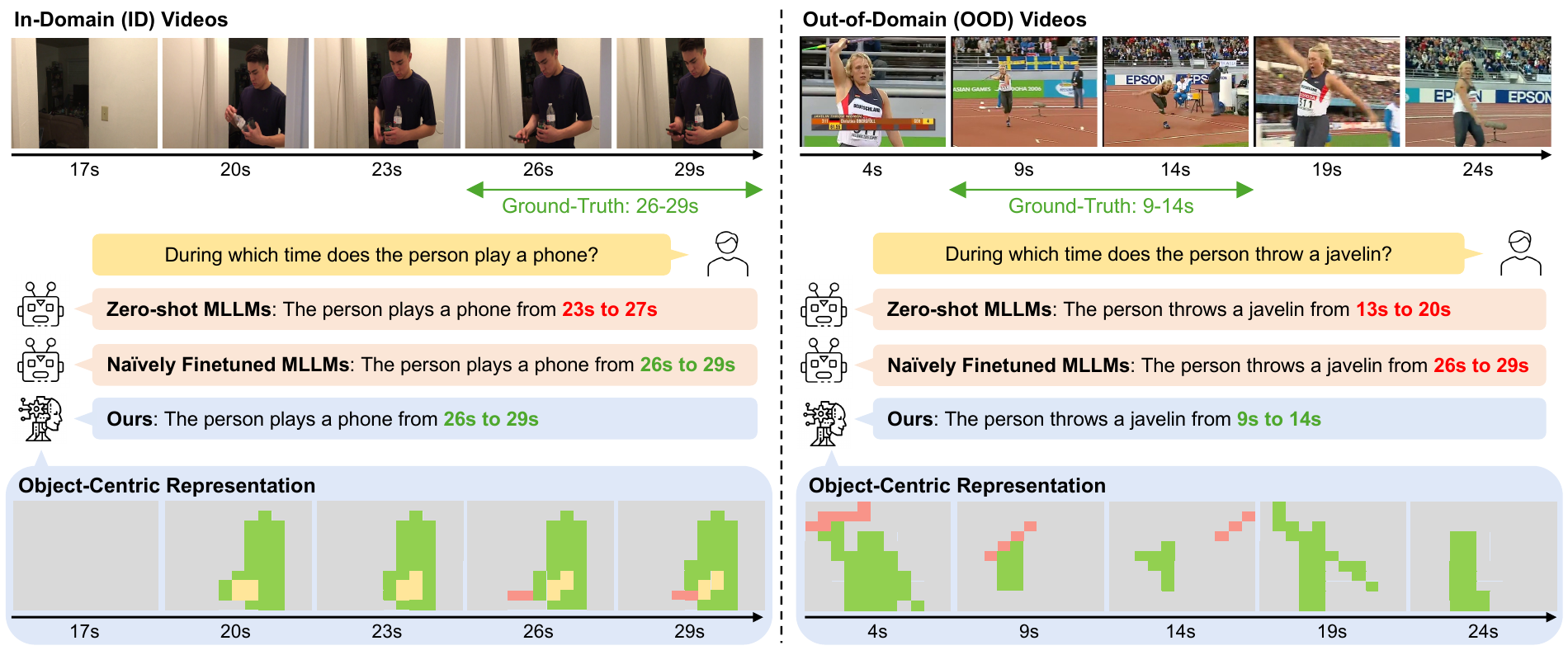}
    \captionof{figure}{\textbf{Motivation.} Zero-shot MLLMs lack fine-grained temporal understanding, producing incorrect timestamps in both settings. Fine-tuning on a VTG dataset resolves this for In-Domain (ID) videos (left), but on Out-of-Domain (OOD) videos the model predicts timestamps based on dataset-specific \emph{\textbf{shortcuts}} rather than the actual visual content (right). Our method leverages \emph{\textbf{object-centric visual representations}} (bottom) that decompose each frame into semantic entities, encouraging genuine visual grounding in both seen and unseen settings.}
\label{fig:teaser}
\end{center}
}]

\let\thefootnote\relax\footnotetext{${}^{*}$Equally contributed first authors. ${}^{\dagger}$Corresponding author. }

\input{0_abstract}    
\input{1_introduction}
\input{2_prior}
\input{2.5_Background}
\input{3_method}

\input{4_result}
\input{5_conclusion}
{
    \small
    \bibliographystyle{ieeenat_fullname}
    \bibliography{main}
}

\end{document}

%% file: preamble.tex
\newcommand{\mpage}[2]
{
\begin{minipage}{#1\linewidth}\centering
#2
\end{minipage}
}

\usepackage{multirow}
\usepackage[table]{xcolor}
\definecolor{bggray}{gray}{0.9}



%% file: 0_abstract.tex
\begin{abstract}

Multimodal Large Language Models (MLLMs) have shown strong performance on Video Temporal Grounding (VTG). However, their coarse recognition capabilities are insufficient for fine-grained temporal understanding, making task-specific fine-tuning indispensable. This fine-tuning causes models to memorize dataset-specific shortcuts rather than faithfully grounding in the actual visual content, leading to poor Out-of-Domain (OOD) generalization. Object-centric learning offers a promising remedy by decomposing scenes into entity-level representations, but existing approaches require re-running the entire multi-stage training pipeline from scratch. We propose SlotVTG, a framework that steers MLLMs toward object-centric, input-grounded visual reasoning at minimal cost. SlotVTG introduces a lightweight slot adapter that decomposes visual tokens into abstract slots via slot attention and reconstructs the original sequence, where objectness priors from a self-supervised vision model encourage semantically coherent slot formation. Cross-domain evaluation on standard VTG benchmarks demonstrates that our approach significantly improves OOD robustness while maintaining competitive In-Domain (ID) performance with minimal overhead.

\end{abstract}

%% file: 1_introduction.tex
\section{Introduction}
\label{sec:intro}

Video Temporal Grounding (VTG), the task of localizing temporal moments in untrimmed videos given natural language queries, has been predominantly addressed by DETR-based specialist models~\cite{lei2021detecting,moon2023query,lin2023univtg,sun2024tr}.
Recently, Multimodal Large Language Models (MLLMs) have emerged as a compelling alternative~\cite{huang2024vtimellm,ren2024timechat,guo2024trace,wang2024hawkeye,meinardus2024chrono,zeng2025timesuite}, owing to their powerful visual representations learned from massive image and video corpora.

However, naively applying MLLMs to VTG yields suboptimal results.
Temporal grounding demands fine-grained temporal understanding that goes beyond the coarse recognition capabilities of general-purpose MLLMs, making task-specific fine-tuning indispensable~\cite{huang2024vtimellm,ren2024timechat,wang2024hawkeye,zeng2025timesuite}.
Yet VTG annotations require precise start-end timestamps for each query, making large-scale data collection prohibitively expensive and preventing models from being exposed to diverse data distributions.
This leads to severe overfitting to dataset-specific shortcuts, as these limited-scale datasets inevitably contain various forms of bias, such as temporal location bias~\cite{chae2024towards,otani2020uncovering,hao2022can}, query text bias~\cite{chae2024towards,li2022compositional,jiang2024mgqp}, and appearance bias~\cite{bao2022debiasttl,qi2024bssard}.
Consequently, these models exhibit severe performance degradation when encountering Out-of-Domain (OOD) test samples (\cref{fig:teaser,fig:analysis}(a)).

In this work, we focus on investigating how the \textit{visual domain gap} leads to VTG performance degradation in OOD settings through comprehensive empirical analyses.
As shown in \cref{fig:analysis}(b), the fine-tuned MLLM exhibits a performance gap of around 13\% on OOD samples, depending on whether they are visually similar or dissimilar to the source dataset.
To further diagnose whether the model genuinely grounds in visual contents, we inject noise into ground-truth segments and compare against perturbing random non-ground-truth segments (\cref{fig:analysis}(c)). On ID, ground-truth perturbation causes a significantly larger drop than random perturbation, confirming the model \emph{does} attend to the target moment. On OOD, however, the two cause nearly identical drops, indicating the model is \emph{not} actively grounding the visual inputs but has rather lost its scene recognition capability for unseen domains.

To encourage the model to genuinely ground on visual contents regardless of domain shifts, it is crucial to extract domain-invariant visual cues.
A promising direction is \emph{object-centric learning}~\cite{locatello2020object}, which decomposes scenes into discrete entity-level representations and has been shown to improve domain generalization in video understanding tasks in MLLMs~\cite{slotvlm,chi2025slot}. However, these approaches integrate object-centric representations between the visual encoder and the language model, requiring the entire vision-language alignment and instruction tuning pipeline to be re-trained from scratch.

We propose \textbf{SlotVTG}, a framework that brings object-centric representation learning into the MLLM framework at minimal cost.
SlotVTG introduces a lightweight \textbf{Slot Adapter} that decomposes visual tokens into a compact set of abstract slots via slot attention, then reconstructs the original token sequence from these slots.
This bottleneck guides visual information through entity-level representations, encouraging the model to suppress spurious correlations and instead attend to the actual visual content relevant to the query.
To further encourage semantically coherent tokens to be grouped into the same slot, we introduce a \textbf{Slot Alignment (SA) loss} that aligns the slot attention maps with self-supervised objectness priors from pre-trained DINOv2~\cite{oquab2023dinov2} features.

We validate our approach through cross-domain evaluation on standard VTG benchmarks, training on one source (\eg, Charades-STA~\cite{gao2017tall} and QVHighlights~\cite{lei2021detecting}) and evaluating on different targets. Our experiments demonstrate that Slot Adapter improves OOD robustness while maintaining competitive ID performance, with minimal memory overhead and additional parameters. 
Our main contributions are as follows:
\begin{itemize}
\item We identify that fine-tuned MLLMs memorize dataset-specific visual shortcuts rather than grounding in the actual visual content.
\item We propose SlotVTG, a parameter-efficient framework consisting of a Slot Adapter that decomposes visual tokens into entity-level slots, and a Slot Alignment Loss that encourages semantically coherent slot formation via objectness priors from pre-trained DINOv2 features.
\item We demonstrate through cross-domain evaluation that SlotVTG significantly improves OOD robustness while maintaining competitive ID performance with minimal overhead.
\end{itemize}

%% file: 2_prior.tex
\begin{figure*}[t]
  \centering
  \includegraphics[width=\linewidth]{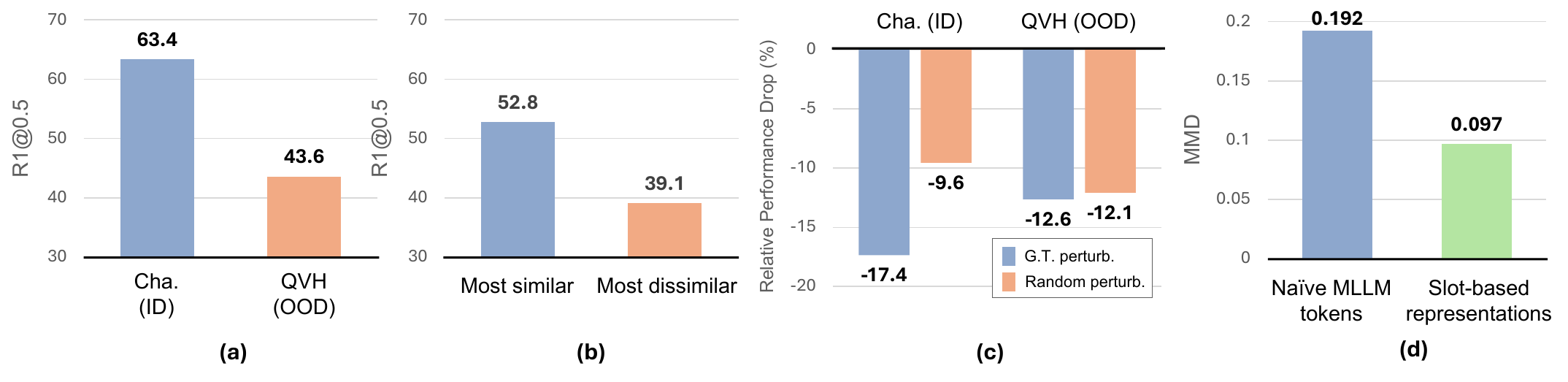}

\caption{\textbf{Observations.} We naively fine-tune Qwen2.5-VL-3B~\cite{bai2025qwen25vl} on Charades-STA (Cha.)~\cite{gao2017tall} (source) and evaluate on QVHighlights (QVH)~\cite{lei2021detecting} (target). \textbf{(a) ID vs.\ OOD performance.}~The model achieves 63.4 R1@0.5 on ID but drops to 43.6 on OOD, confirming severe overfitting to dataset-specific patterns. \textbf{(b) Visual similarity analysis.}~We extract visual features from the vision encoder and compute cosine similarity between ID and OOD samples; performance on the most similar 20\% of OOD samples (52.8) far exceeds the most dissimilar 20\% (39.1), indicating that the model fails when visual distribution shifts. \textbf{(c) Noise perturbation.}~We report R1@0.7 for a stricter localization threshold. On ID, ground-truth perturbation causes a 17.4\% drop while random perturbation causes only 9.6\%, a significant gap confirming the model attends to the target moment. On OOD, however, the two cause nearly identical drops (12.6\% vs.\ 12.1\%), revealing that the model does not attend to the actual visual content under distribution shift. \textbf{(d) Domain gap.}~MMD distance~\cite{gretton2012kernel} of our slot-based representations (0.097) is substantially lower than the baseline (0.192), showing that object-centric decomposition reduces the domain gap between source and target distributions.}
  \label{fig:analysis}
\end{figure*}

\section{Related Work}
\label{sec:related}

\subsection{Video Temporal Grounding}
Video Temporal Grounding (VTG) aims to localize temporal moments in untrimmed videos given natural language queries. Early approaches rely on proposal-based or regression-based architectures~\cite{gao2017tall,anne2017localizing,zhang2020learning,zeng2020dense}. Inspired by the success of DETR~\cite{carion2020end} in object detection, Moment-DETR~\cite{lei2021detecting} pioneered the use of set prediction for joint moment retrieval and highlight detection, establishing the QVHighlights benchmark. Subsequent DETR-based methods have advanced the paradigm through query-dependent representations~\cite{moon2023query}, event-aware attention~\cite{jang2023knowing}, unified multi-task frameworks~\cite{lin2023univtg}, task-reciprocal decoding~\cite{sun2024tr}, correlation-guided calibration~\cite{moon2023correlation}, and joint task exploration~\cite{yang2024task,xiao2024bridging}.

More recently, Multimodal Large Language Models (MLLMs) have emerged as a compelling alternative. VTimeLLM~\cite{huang2024vtimellm} and TimeChat~\cite{ren2024timechat} demonstrate that MLLMs can generate temporal boundaries as text tokens through task-specific instruction tuning. This generative paradigm has been extended by interleaved frame-timestamp representations~\cite{meinardus2024chrono}, causal event modeling~\cite{guo2024trace}, grounded tuning for long videos~\cite{zeng2025timesuite}, chain-of-LoRA reasoning~\cite{liu2026videomind}, and reinforcement learning with verifiable temporal rewards~\cite{wang2025timer}. While these methods improve temporal understanding within MLLMs, they focus on architectural and training innovations without addressing the fundamental problem of dataset-specific shortcut learning during fine-tuning.

\subsection{Bias in Video Understanding}
Dataset bias has been widely studied across video understanding tasks. In action recognition, Choi~\etal~\cite{choi2019can} reveal that models exploit scene context as a shortcut, achieving high accuracy without attending to the actual action. Li~\etal~\cite{li2018resound} formalize representation bias in video datasets and introduce the Diving48 benchmark to mitigate it. Bae~\etal~\cite{bae2024devias} further address this through disentangled action-scene representations. Beyond action recognition, Lei~\etal~\cite{lei2023revealing} show that single-frame models perform surprisingly well on video-language tasks, exposing static appearance bias.

In the VTG domain specifically, Otani~\etal~\cite{otani2020uncovering} demonstrate that blind baselines without video input can match trained models by exploiting annotation distribution patterns. This finding prompted the creation of out-of-distribution evaluation splits~\cite{yuan2021closer} and spurred numerous debiasing methods. Causal inference approaches~\cite{yang2021deconfounded,nan2021interventional} use backdoor adjustment to remove confounding effects of moment location. Adversarial and augmentation strategies~\cite{qi2024bssard,hao2022can,lan2023curriculum} synthesize bias-conflict samples or shuffle temporal structure to discourage shortcut exploitation. Chae~\etal~\cite{chae2024towards} provide a comprehensive benchmark across seven datasets, analyzing annotation bias and query text patterns. These studies collectively reveal that VTG datasets contain diverse biases spanning annotation distributions, language patterns, and visual modalities, and that existing models remain vulnerable to exploiting such shortcuts rather than performing genuine cross-modal grounding.

\subsection{Object-Centric Learning}
Object-centric learning aims to decompose scenes into discrete entity-level representations. Slot Attention~\cite{locatello2020object} introduces an iterative competitive attention mechanism where learnable slots compete to explain input tokens. DINOSAUR~\cite{seitzer2022bridging} extends slot attention to real-world images by reconstructing self-supervised DINO~\cite{dino} features instead of raw pixels. In the video domain, SAVi~\cite{kipf2021conditional} conditions slot initialization on optical flow for temporal consistency, while SlotFormer~\cite{wu2022slotformer} learns unsupervised visual dynamics through autoregressive slot prediction.

Integrating object-centric representations into vision-language models is an emerging direction. Slot-VLM~\cite{slotvlm} designs dual-branch object-event slots that decompose video tokens into object-centric and event-centric representations for LLM reasoning. Slot-MLLM~\cite{chi2025slot} combines Q-Former with slot attention to produce discrete object-centric visual tokens for unified multimodal generation. However, both approaches require training the entire vision-language pipeline from scratch, including visual token alignment and instruction tuning. Our work differs in that we introduce a lightweight slot-based adapter that can be attached to existing fine-tuned MLLMs at minimal cost, without modifying the base training pipeline.

%% file: 2.5_Background.tex
\section{Preliminaries}
\label{sec:background}

\begin{figure*}[t]
\centering
\includegraphics[width=\textwidth]{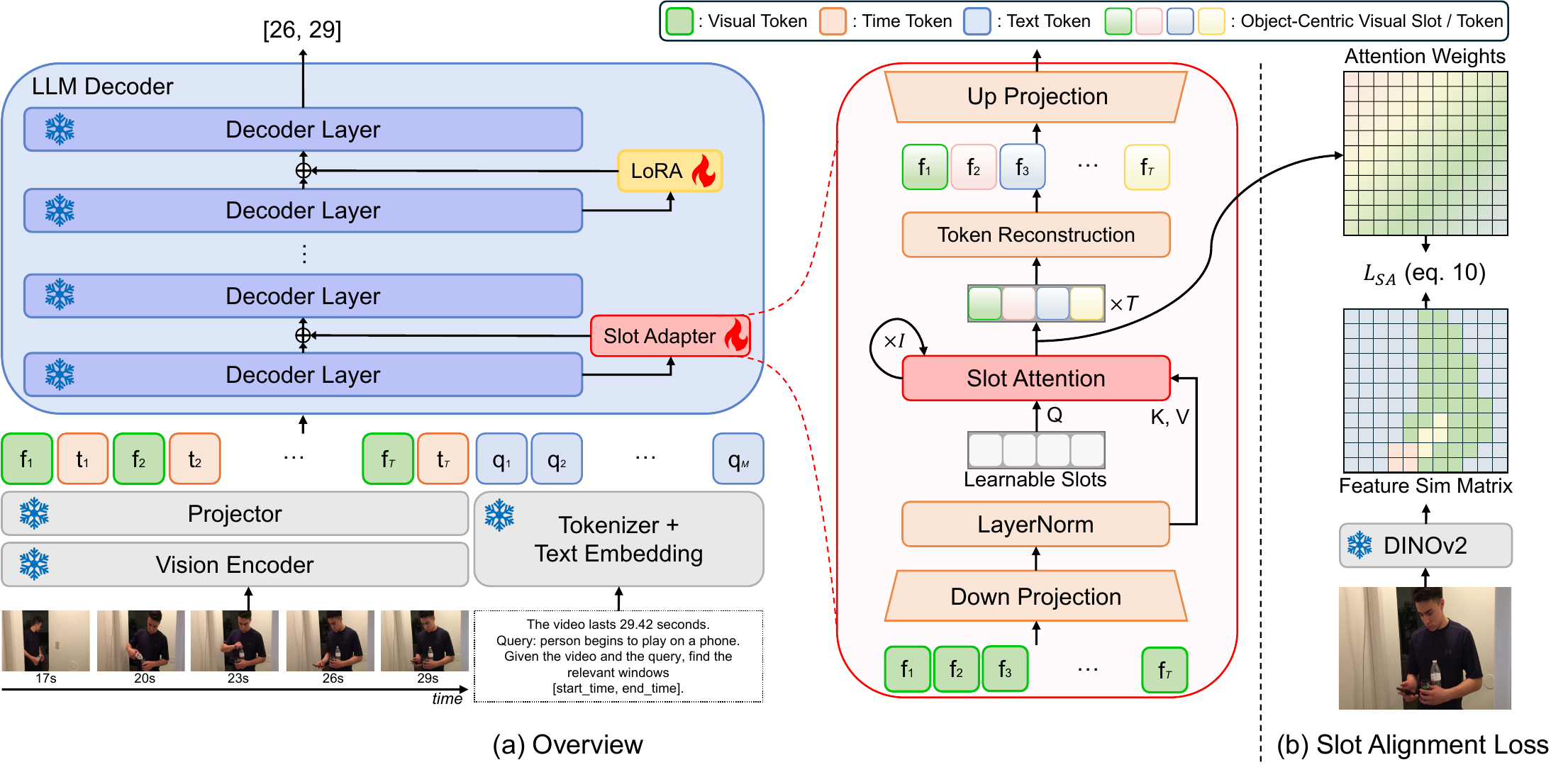}
\caption{\textbf{(a) Overview of SlotVTG.} Video frames are encoded into visual tokens and projected into the LLM decoder. In the early decoder layers, a lightweight \emph{Slot Adapter} decomposes visual tokens into entity-level slots via iterative slot attention, then reconstructs the token sequence. The resulting tokens carry disentangled, entity-aware representations before entering the later layers, which are fine-tuned with LoRA for temporal reasoning and answer generation. Text tokens bypass the Slot Adapter throughout. \textbf{(b) Slot Alignment Loss.} Token-pair similarity derived from slot attention weights is aligned with that from a pre-trained DINOv2 model, encouraging semantically coherent tokens to be grouped into the same slot.}
\label{fig:main}
\end{figure*}

\subsection{Generative VTG Framework}
We follow the generative Video Temporal Grounding (VTG) paradigm, where an MLLM directly generates target timestamps as text tokens.
Given $T$ uniformly sampled video frames and a natural language query, we encode each frame into $N$ visual tokens $\mathbf{f}_i \in \mathbb{R}^{N \times D}$ via a frozen vision encoder and linear projection, where $D$ is the hidden dimension of the LLM decoder, and tokenize its timestamp into a short text sequence $\mathbf{t}_i$.
Each frame's sampling time in seconds is tokenized into a short text sequence $\mathbf{t}_i$ (e.g., ``2.5s'').
The input to the LLM decoder is constructed by interleaving each frame's visual tokens with its timestamp tokens, followed by the query tokens $\mathbf{q}$:
\begin{equation}
    \mathbf{x} = [\mathbf{f}_1, \mathbf{t}_1, \mathbf{f}_2, \mathbf{t}_2, \dots, \mathbf{f}_T, \mathbf{t}_T, \mathbf{q}],
    \label{eq:input_sequence}
\end{equation}
This interleaved layout has been shown to be effective for temporal grounding~\cite{meinardus2024chrono, zhang2025timelens}.
The model autoregressively decodes the target temporal window $[t_{\text{start}}, t_{\text{end}}]$.

\subsection{Observations}
\label{sec:failure_analysis}

To understand why na\"ively fine-tuned MLLMs fail under distribution shift, we conduct a series of diagnostic experiments. We fine-tune Qwen2.5-VL-3B on Charades-STA~\cite{gao2017tall} and evaluate on QVHighlights~\cite{lei2021detecting}.

\vspace{2mm}
\noindent\textbf{OOD performance degradation.}
\cref{fig:analysis}(a) compares ID and OOD performance. The fine-tuned model achieves 63.4 R1@0.5 on ID but only 43.6 on OOD---a 31.2\% relative drop. This confirms that the model overfits to source-domain patterns rather than learning generalizable temporal grounding.

\vspace{2mm}
\noindent\textbf{Visual similarity matters.}
To investigate whether this degradation correlates with visual distribution shift, we extract features from the vision encoder and rank OOD samples by cosine similarity to the training set. As shown in \cref{fig:analysis}(b), the most similar 20\% of OOD samples achieve 52.8 R1@0.5, while the most dissimilar 20\% drop to 39.1. This reveals that the model's predictions degrade proportionally with visual domain distance, suggesting it relies on surface-level visual patterns seen during training.

\vspace{2mm}
\noindent\textbf{The model ignores visual content on OOD.}
We design a noise perturbation experiment to directly test whether the model attends to the visual content within ground-truth segments. Specifically, we add Gaussian noise to the visual tokens corresponding to the annotated temporal window and measure the performance change. We report R1@0.7 for this experiment, as a stricter IoU threshold better captures whether the model precisely localizes the target moment. As shown in \cref{fig:analysis}(c), corrupting the ground-truth segment on ID causes a 17.4\% drop, while corrupting random non-ground-truth segments causes only a 9.6\% drop---a 7.8\%p gap confirming the model \emph{does} rely on the target moment. On OOD, however, ground-truth perturbation (12.6\%) and random perturbation (12.1\%) cause nearly identical degradation, with only a 0.5\%p gap---the model is effectively ignoring the visual content of the target moment and instead relying on dataset-specific shortcuts.

\vspace{2mm}
\noindent\textbf{Object-centric representations reduce domain gap.}
The above findings motivate our approach: if the model fails because it relies on domain-specific visual patterns, decomposing the representation into object-centric slots should yield more transferable features. \cref{fig:analysis}(d) validates this hypothesis. We compute a per-video representation by averaging the vision token hidden states and measure the Maximum Mean Discrepancy (MMD)~\cite{gretton2012kernel} between source and target distributions (see \cref{sec:imple_detail} for details). 
The baseline exhibits an MMD of 0.192, while our slot-based representation reduces it to 0.097 (-49.6\%), demonstrating that object-centric decomposition substantially narrows the domain gap.

%% file: 3_method.tex
\section{SlotVTG}
\label{sec:method}

We introduce \textbf{SlotVTG}, a parameter-efficient framework that brings object-centric visual representation into pre-trained MLLMs at minimal cost.
\cref{fig:main}(a) provides an overview.
We describe the Slot Adapter in \cref{sec:slot_adapter}, the Slot Alignment Loss in \cref{sec:sa_loss}, and the training objective in \cref{sec:loss}.

\subsection{Slot Adapter}
\label{sec:slot_adapter}
Let $\mathbf{X} \in \mathbb{R}^{T \times N \times D}$ denote the visual tokens at a given decoder layer, where $N$ is the number of tokens per frame and $D$ is the hidden dimension.
Instead of letting the LLM decoder process these tokens directly, we decompose them into a compact set of $N_s$ abstract slots via iterative slot attention~\cite{locatello2020object}.

\noindent\textbf{Down Projection.}
We first project the visual tokens $\mathbf{X}$ into a lower-dimensional bottleneck space:
\begin{equation}
\mathbf{X}_{down} = \mathbf{X}\mathbf{W}_{down} \in \mathbb{R}^{T \times N \times d}
\end{equation}
where $\mathbf{W}_{down} \in \mathbb{R}^{D \times d}$ and $d \ll D$.

\noindent\textbf{Slot Attention.}
A set of $N_s$ learnable slot queries $\mathbf{S}^{(0)} \in \mathbb{R}^{T \times N_s \times d}$ attend to the projected tokens through $I$ iterations.
At each iteration, we project the slots and tokens into a common space with dimension $d_h$: $\mathbf{Q} = \mathbf{S}^{(i)} \mathbf{W}_Q \in \mathbb{R}^{T \times N_s \times d_h}$, $\mathbf{K} = \mathbf{X}_{down} \mathbf{W}_K \in \mathbb{R}^{T \times N \times d_h}$, and $\mathbf{V} = \mathbf{X}_{down} \mathbf{W}_V \in \mathbb{R}^{T \times N \times d_h}$, where $\mathbf{W}_Q$, $\mathbf{W}_K$, and $\mathbf{W}_V$ are the projection matrices.
The attention scores are computed as:
\begin{equation}
\mathbf{M} = \mathbf{K} \mathbf{Q}^T / \sqrt{d_h} \in \mathbb{R}^{T \times N \times N_s}
\end{equation}
We normalize $\mathbf{M}$ along the \emph{slot axis} via softmax, fostering competitive assignment of tokens to slots:
\begin{equation}
A(n, k) = \frac{\exp(M(n, k))}{\sum_{j=1}^{N_s} \exp(M(n, j))}
\end{equation}
We then normalize $\mathbf{A}$ along the token axis such that $\hat{A}(\cdot, k)$ sums to one:
\begin{equation}
\hat{A}(n, k) = \frac{A(n, k)}{\sum_{j=1}^{N} A(j, k)}
\end{equation}
The updated slot representations are computed as a weighted mean aggregation $\mathbf{Z} = \hat{\mathbf{A}}^T \mathbf{V}$.
Slots are updated via a Gated Recurrent Unit (GRU)~\cite{cho2014gru} based recurrence.
This competition mechanism encourages each slot to specialize in a distinct semantic entity within the frame.

\noindent\textbf{Token Reconstruction.}
Since the LLM decoder expects the original token sequence length, we reconstruct the visual tokens from the slots via cross-attention, where the original tokens act as queries to retrieve entity-aware information from the final slots $\mathbf{S}^{(I)}$:
\begin{equation}
\hat{\mathbf{X}} = \text{CrossAttn}(\mathbf{X}_{down},\ \mathbf{S}^{(I)}) \in \mathbb{R}^{T \times N \times d}
\end{equation}
The reconstructed tokens are projected back to the original dimension via an up projection.
The adapter output is then added to the original tokens via a residual connection with a zero-initialized projection:
\begin{equation}
\mathbf{X}_{out} = \mathbf{X} + \hat{\mathbf{X}}\mathbf{W}_{up}
\end{equation}
where $\mathbf{W}_{up} \in \mathbb{R}^{d \times D}$ projects back to the original dimension and is initialized to zero, so the adapter acts as an identity mapping at the start of training.
This ensures training stability while gradually steering the representations toward entity-level decomposition.

\noindent\textbf{Early-Layer Insertion.}
We attach the Slot Adapter only to the early decoder layers.
Recent findings~\cite{maptheflow} show that cross-frame interactions occur in these early layers, while deeper layers handle language integration and answer generation.
By inserting the Slot Adapter at this stage, each slot captures temporally coherent semantics across frames rather than frame-independent decompositions.
The deeper layers, fine-tuned with LoRA~\cite{hu2022lora,yang2023aim}, then reason over these disentangled representations.
Text tokens bypass the Slot Adapter throughout.

\input{table/results}

\subsection{Slot Alignment Loss}
\label{sec:sa_loss}
While the Slot Adapter encourages decomposition through its bottleneck structure, the slots may form arbitrary clusters without additional guidance.
We {introduce Slot Alignment (SA) loss, which} distills objectness priors from a self-supervised vision model (DINOv2~\cite{oquab2023dinov2}) to encourage semantically coherent slot formation, as illustrated in \cref{fig:main}(b).

\noindent\textbf{Slot-based Similarity.}
Let $\mathbf{A} \in \mathbb{R}^{T \times N_s \times N}$ denote the slot attention weights from the final iteration.
We transpose and $L_2$-normalize along the slot dimension, yielding $\bar{\mathbf{A}} \in \mathbb{R}^{T \times N \times N_s}$.
Token-pair similarity under the slot assignments is computed and rescaled to $[-1, 1]$ to match the range of cosine similarity:
\begin{equation}
\mathbf{M}_{slot} = 2(\bar{\mathbf{A}}\bar{\mathbf{A}}^T) - 1 \in \mathbb{R}^{T \times N \times N}
\end{equation}

\noindent\textbf{DINO-based Similarity.}
We extract features from the last transformer block of a pre-trained DINOv2~\cite{oquab2023dinov2} model and $L_2$-normalize them, yielding $\bar{\mathbf{F}}_{dino} \in \mathbb{R}^{T \times N \times d_{dino}}$.
The target token-pair similarity is then computed as:
\begin{equation}
\mathbf{M}_{dino} = \bar{\mathbf{F}}_{dino}\bar{\mathbf{F}}_{dino}^T \in \mathbb{R}^{T \times N \times N}
\end{equation}

\noindent\textbf{Loss.}
The {SA} loss 
aligns these two structures:
\begin{equation}
\mathcal{L}_{SA} = 1 - \frac{1}{T}\sum_{t=1}^{T}\cos\left((\mathbf{M}_{slot}^{(t)}),\ (\mathbf{M}_{dino}^{(t)})\right)
\end{equation}


\subsection{Training Objective}
\label{sec:loss}
The framework is trained end-to-end with the vision encoder frozen. The Slot Adapters and LoRA parameters are updated jointly. The total loss combines the standard autoregressive cross-entropy loss with the slot alignment regularization:
\begin{equation}
\mathcal{L}_{total} = \mathcal{L}_{CE} + \lambda\mathcal{L}_{SA}
\end{equation}
where $\lambda$ controls the strength of the objectness prior.

%% file: table/results.tex
\begin{table*}[t]
\centering
\caption{\textbf{Performance comparison on video temporal grounding benchmarks.} 
We evaluate SlotVTG against state-of-the-art models on Charades-STA~\cite{gao2017tall}, QVHighlights~\cite{lei2021detecting}, and ActivityNet Captions~\cite{krishna2017dense}.
We report both \textcolor{gray}{In-Domain (ID)} settings, where the source and target datasets are the same, and Out-of-Distribution (OOD) settings, where they differ.
DETR-based methods (EATR~\cite{jang2023knowing} and CG-DETR~\cite{moon2023correlation}) are reproduced using pre-extracted CLIP~\cite{radford2021learning} + SlowFast~\cite{feichtenhofer2019slowfast} features at 0.5 fps, following their original implementation.
The performance of \colorbox{bggray}{zero-shot} VTG models is reported for reference.
\colorbox{green!10}{Our results} are highlighted in green. The best results under the same cross-domain evaluation setting (source $\rightarrow$ target, LLM size) are highlighted in \textbf{bold}.
}

\renewcommand{\arraystretch}{1.3}
\resizebox{\textwidth}{!}{
\begin{tabular}{c l c cccc c cccc c cccc}
\toprule
\multirow{3}{*}{\shortstack{Source\\dataset}} & \multirow{3}{*}{Method} & \multirow{3}{*}{\shortstack{LLM\\size}} & \multicolumn{14}{c}{Target dataset} \\
\cmidrule{4-17}
& & & \multicolumn{4}{c}{Charades-STA} && \multicolumn{4}{c}{ActivityNet-Captions} && \multicolumn{4}{c}{QVHighlights} \\
\cmidrule{4-7} \cmidrule{9-12} \cmidrule{14-17}
& & & R1@0.3 & R1@0.5 & R1@0.7 & mIoU && R1@0.3 & R1@0.5 & R1@0.7 & mIoU && R1@0.3 & R1@0.5 & R1@0.7 & mIoU \\
\midrule
\multirow{5}{*}{\rotatebox[origin=c]{90}{Zero-Shot}} 
& \cellcolor{bggray} HawkEye ~\cite{wang2024hawkeye} & \cellcolor{bggray}7B & \cellcolor{bggray}50.6 & \cellcolor{bggray}31.4 & \cellcolor{bggray}14.5 & \cellcolor{bggray}33.7 &\cellcolor{bggray}& \cellcolor{bggray}49.1 & \cellcolor{bggray}29.3 & \cellcolor{bggray}10.7 & \cellcolor{bggray}32.7 &\cellcolor{bggray}& \cellcolor{bggray}- & \cellcolor{bggray}- & \cellcolor{bggray}- & \cellcolor{bggray}-\\
& \cellcolor{bggray} TimeSuite ~\cite{zeng2025timesuite} & \cellcolor{bggray}7B & \cellcolor{bggray}69.9 & \cellcolor{bggray}48.7 & \cellcolor{bggray}24.0 & \cellcolor{bggray}- &\cellcolor{bggray}& \cellcolor{bggray}- & \cellcolor{bggray}16.6 & \cellcolor{bggray}9.3 & \cellcolor{bggray}22.0 &\cellcolor{bggray}& \cellcolor{bggray}-  & \cellcolor{bggray}12.3 & \cellcolor{bggray}9.2 & \cellcolor{bggray}21.3 \\
& \cellcolor{bggray}UniTime ~\cite{li2025unitime} & \cellcolor{bggray}7B & \cellcolor{bggray}- & \cellcolor{bggray}{59.1} & \cellcolor{bggray}{31.9} & \cellcolor{bggray}{52.2} &\cellcolor{bggray}& \cellcolor{bggray}- & \cellcolor{bggray}22.8 & \cellcolor{bggray}14.1 & \cellcolor{bggray}27.3 &\cellcolor{bggray}& \cellcolor{bggray}- & \cellcolor{bggray}41.0 & \cellcolor{bggray}31.5 & \cellcolor{bggray}43.7 \\
& \cellcolor{bggray} VideoMind ~\cite{liu2026videomind} & \cellcolor{bggray}2B & \cellcolor{bggray}67.6 & \cellcolor{bggray}51.1 & \cellcolor{bggray}26.0 & \cellcolor{bggray}45.2 &\cellcolor{bggray}& \cellcolor{bggray}44.0 & \cellcolor{bggray}26.5 & \cellcolor{bggray}12.6 & \cellcolor{bggray}30.1 &\cellcolor{bggray}& \cellcolor{bggray}- & \cellcolor{bggray}- & \cellcolor{bggray}- & \cellcolor{bggray}- \\
& \cellcolor{bggray} VideoMind ~\cite{liu2026videomind} & \cellcolor{bggray}7B & \cellcolor{bggray}73.5 & \cellcolor{bggray}{59.1} & \cellcolor{bggray}31.2 & \cellcolor{bggray}50.2 &\cellcolor{bggray}& \cellcolor{bggray}48.4 & \cellcolor{bggray}30.3 & \cellcolor{bggray}15.7 & \cellcolor{bggray}33.3 &\cellcolor{bggray}& \cellcolor{bggray}- & \cellcolor{bggray}- & \cellcolor{bggray}- & \cellcolor{bggray}- \\
\midrule
\multirow{7}{*}{\rotatebox[origin=c]{90}{Charades-STA}} 
& EaTR~\cite{jang2023knowing} & - & \textcolor{gray}{67.7} & \textcolor{gray}{55.2} & \textcolor{gray}{33.1} & \textcolor{gray}{47.7} && 36.9 & 18.8 & 7.3 & 24.1 && 31.7 & 17.0 & 6.4 & 21.5 \\
& CG-DETR~\cite{moon2023correlation} & - & \textcolor{gray}{69.7} & \textcolor{gray}{57.6} & \textcolor{gray}{35.1} & \textcolor{gray}{49.5} && 32.6 & 16.8 & 6.8 & 22.1 && 37.4 & 22.8 & 10.5 & 25.2 \\
& Chrono-BLIP~\cite{meinardus2024chrono}  & 4B & \textcolor{gray}{77.5} & \textcolor{gray}{68.8} & \textcolor{gray}{48.5} & \textcolor{gray}{57.2} && 41.8 & 22.4 & 9.7 & 27.7 && \textbf{66.6} & 43.9 & 23.7 & 43.9 \\
& Chrono-Qwen~\cite{meinardus2024chrono}  & 3B & \textcolor{gray}{77.2} & \textcolor{gray}{63.4} & \textcolor{gray}{40.3} & \textcolor{gray}{55.2} && 44.4 & 26.3 & 13.1 & 30.1 && 63.3 & 43.6 & 23.3 & 42.7 \\
& \cellcolor{green!10}SlotVTG (Ours) & \cellcolor{green!10}3B & \cellcolor{green!10}\textcolor{gray}{77.2} & \cellcolor{green!10}\textcolor{gray}{64.0} & \cellcolor{green!10}\textcolor{gray}{41.2} & \cellcolor{green!10}\textcolor{gray}{55.4} &\cellcolor{green!10}& \cellcolor{green!10}\textbf{47.7} & \cellcolor{green!10}\textbf{28.7} & \cellcolor{green!10}\textbf{14.4} & \cellcolor{green!10}\textbf{32.2} &\cellcolor{green!10}& \cellcolor{green!10}66.0 & \cellcolor{green!10}\textbf{47.9} & \cellcolor{green!10}\textbf{26.2} & \cellcolor{green!10}\textbf{45.0} \\
& Chrono-Qwen~\cite{meinardus2024chrono}  & 7B & \textcolor{gray}{79.1} & \textcolor{gray}{67.8} & \textcolor{gray}{46.9} & \textcolor{gray}{58.1} && 46.5 & 29.2 & 14.6 & 32.6 && 70.3 & 53.5 & 29.6 & 49.0 \\
& \cellcolor{green!10}SlotVTG (Ours) & \cellcolor{green!10}7B & \cellcolor{green!10}\textcolor{gray}{79.5} & \cellcolor{green!10}\textcolor{gray}{67.6} & \cellcolor{green!10}\textcolor{gray}{46.7} & \cellcolor{green!10}\textcolor{gray}{58.3} &\cellcolor{green!10}& \cellcolor{green!10}\textbf{52.0} & \cellcolor{green!10}\textbf{33.2} & \cellcolor{green!10}\textbf{16.7} & \cellcolor{green!10}\textbf{35.5} &\cellcolor{green!10}& \cellcolor{green!10}\textbf{74.0} & \cellcolor{green!10}\textbf{57.6} & \cellcolor{green!10}\textbf{32.2} & \cellcolor{green!10}\textbf{51.3} \\
\midrule
\multirow{7}{*}{\rotatebox[origin=c]{90}{QVHighlights}} 
& EaTR~\cite{jang2023knowing} & - & 40.8 & 27.2 & 13.0 & 28.0 && 36.7 & 20.9 & 9.7 & 25.3 && \textcolor{gray}{70.3} & \textcolor{gray}{59.6} & \textcolor{gray}{40.3} & \textcolor{gray}{53.1} \\
& CG-DETR~\cite{moon2023correlation} & - & 42.8 & 25.5 & 12.2 & 28.5 && 37.7 & 21.5 & 10.4 & 26.0 && \textcolor{gray}{77.5} & \textcolor{gray}{65.6} & \textcolor{gray}{52.1} & \textcolor{gray}{61.3} \\
& Chrono-BLIP~\cite{meinardus2024chrono}  & 4B & 61.5 & 37.0 & 19.8 & 41.4 && 41.8 & 22.4 & 9.7 & 27.7 && \textcolor{gray}{86.1} & \textcolor{gray}{76.8} & \textcolor{gray}{62.8} & \textcolor{gray}{70.8} \\
& Chrono-Qwen~\cite{meinardus2024chrono}  & 3B & 70.6 & 45.7 & 21.8 & 45.7 && 55.2 & 35.3 & 20.8 & 39.2 && \textcolor{gray}{87.6} & \textcolor{gray}{79.1} & \textcolor{gray}{64.8} & \textcolor{gray}{71.7} \\
& \cellcolor{green!10}SlotVTG (Ours) & \cellcolor{green!10}3B & \cellcolor{green!10}\textbf{70.7} & \cellcolor{green!10}\textbf{46.6} & \cellcolor{green!10}\textbf{22.6} & \cellcolor{green!10}\textbf{46.1} &\cellcolor{green!10}& \cellcolor{green!10}\textbf{56.1} & \cellcolor{green!10}\textbf{35.7} & \cellcolor{green!10}\textbf{21.1} & \cellcolor{green!10}\textbf{40.0} &\cellcolor{green!10}& \cellcolor{green!10}\textcolor{gray}{87.3} & \cellcolor{green!10}\textcolor{gray}{79.5} & \cellcolor{green!10}\textcolor{gray}{64.6} & \cellcolor{green!10}\textcolor{gray}{71.7} \\
& Chrono-Qwen~\cite{meinardus2024chrono}  & 7B & 75.2 & 53.3 & 27.4 & 49.9 && 60.7 & 41.4 & 24.8 & 43.4 && \textcolor{gray}{90.7} & \textcolor{gray}{81.8} & \textcolor{gray}{67.6} & \textcolor{gray}{74.9} \\
& \cellcolor{green!10}SlotVTG (Ours) & \cellcolor{green!10}7B & \cellcolor{green!10}\textbf{76.0} & \cellcolor{green!10}\textbf{53.7} & \cellcolor{green!10}\textbf{28.2} & \cellcolor{green!10}\textbf{50.4} &\cellcolor{green!10}& \cellcolor{green!10}\textbf{61.7} & \cellcolor{green!10}\textbf{42.0} & \cellcolor{green!10}\textbf{25.3} & \cellcolor{green!10}\textbf{44.1} &\cellcolor{green!10}& \cellcolor{green!10}\textcolor{gray}{91.3} & \cellcolor{green!10}\textcolor{gray}{82.9} & \cellcolor{green!10}\textcolor{gray}{69.3} & \cellcolor{green!10}\textcolor{gray}{76.0} \\

\bottomrule
\end{tabular}
\label{tab:main}
}
\end{table*}


%% file: 4_result.tex
\section{Experiments}

\subsection{Experimental Setup}
\noindent\textbf{Implementation Details.}
\label{sec:imple_detail}
We build upon Qwen2.5-VL-Instruct~\cite{bai2025qwen25vl} (3B and 7B) as the backbone MLLM, where the LLM decoder has a hidden dimension of $D{=}2048$ and $D{=}3584$, respectively. The vision encoder processes each $224{\times}224$ frame into $N{=}64$ visual tokens ($8{\times}8$ spatial grid).
For the Slot Adapter, we set the bottleneck dimension to $d{=}512$, the number of slots to $K{=}4$, and use 8 attention heads with $I{=}3$ iterations of GRU-based refinement. 
We inject the Slot Adapter into layers 1--7, while LoRA~\cite{hu2022lora} (rank 16, $\alpha{=}64$) is applied to the remaining deeper layers. 
The visual token hidden states used for the MMD analysis in \cref{sec:failure_analysis} are extracted from the last adapter layer (layer 7).
The Slot Alignment loss uses DINOv2~\cite{oquab2023dinov2}-base affinity matrices and is applied at the last layer where the Slot Adapter is inserted (layer 7) with $\lambda{=}0.1$. 
For video processing, we uniformly sample 20 and 60 frames for the models trained on Charades-STA~\cite{gao2017tall} and QVHighlights~\cite{lei2021detecting}, respectively.
We train for 5 epochs with AdamW~\cite{adamw} (learning rate $5{\times}10^{-5}$) and a global batch size of 32 on 8 NVIDIA 3090/4090 GPUs. 
In total, the trainable parameters (Slot Adapters + LoRA) amount to approximately 7.6M (0.25\% of the total) for the 3B model and 23.3M (0.33\% of the total) for the 7B model.

\noindent\textbf{Evaluation Protocol.}
We use Charades-STA (Cha.)~\cite{gao2017tall} and QVHighlights (QVH.)~\cite{lei2021detecting} as source datasets for fine-tuning, and evaluate on three target datasets: Cha., QVH., and ActivityNet Captions (ANet)~\cite{krishna2017dense}. We denote each setting by its source-target pair (\eg, Cha.$\rightarrow$ANet). For each pair, we report both ID performance (source = target) and OOD performance (source $\neq$ target). All results are reported using standard moment retrieval metrics: R1@0.3, R1@0.5, R1@0.7, and mIoU.

\noindent\textbf{Baselines.}
We compare against three categories of methods.
(1)~{Zero-shot MLLMs} that perform VTG without task-specific fine-tuning: HawkEye~\cite{wang2024hawkeye}, TimeSuite~\cite{zeng2025timesuite}, UniTime~\cite{li2025unitime}, and VideoMind~\cite{liu2026videomind}.
(2)~{DETR-based specialists} trained on a single source dataset: EaTR~\cite{jang2023knowing} and CG-DETR~\cite{moon2023correlation}.
(3)~{MLLM-based methods} fine-tuned on a single source dataset: Chrono~\cite{meinardus2024chrono} with both BLIP-2~\cite{li2023blip2} and Qwen2.5-VL-Instruct~\cite{bai2025qwen25vl} (3B and 7B) backbones.

\subsection{Results}

\noindent\textbf{Comparison with State-of-the-Art.}
\cref{tab:main} summarizes the results.
SlotVTG consistently improves OOD performance across all source-target configurations while maintaining competitive ID performance.

When trained on Cha., SlotVTG (3B) achieves substantial OOD gains over the Chrono-Qwen~\cite{meinardus2024chrono} baseline: +2.4 R1@0.5 on ANet and +4.3 R1@0.5 on QVH., while preserving ID performance on Cha.\ (64.0 vs.\ 63.4 R1@0.5).
This trend scales to the 7B model, where OOD improvements are even more pronounced (+4.0 R1@0.5 on ANet and +4.1 R1@0.5 on QVH.), demonstrating that SlotVTG benefits from larger model capacity.

When trained on QVH., SlotVTG (3B) again improves OOD generalization to both Cha.\ (+0.9 R1@0.5) and ANet (+0.4 R1@0.5) without sacrificing ID performance.
SlotVTG (7B) also achieves OOD gains over the baseline: +0.4 R1@0.5 on Cha. and +0.6 R1@0.5 on ANet.
The smaller OOD gains in this setting are expected, as QVH.\ is a more diverse dataset with broader domain coverage, leaving less room for improvement.

Notably, SlotVTG with a 3B backbone trained on Cha.\ already surpasses several zero-shot 7B models in OOD settings (\eg, 28.7 vs.\ 30.3 R1@0.5 on ANet for SlotVTG 3B vs.\ VideoMind~\cite{liu2026videomind} 7B), despite being fine-tuned on a single source dataset.
Compared to DETR-based specialists (EaTR~\cite{jang2023knowing}, CG-DETR~\cite{moon2023correlation}), SlotVTG achieves significantly better OOD performance across all settings.
These results highlight that object-centric decomposition enables the model to genuinely ground in visual content rather than relying on dataset-specific patterns, resulting in robust generalization across domains.

\begin{figure}[t]
\centering
\includegraphics[width=\linewidth]{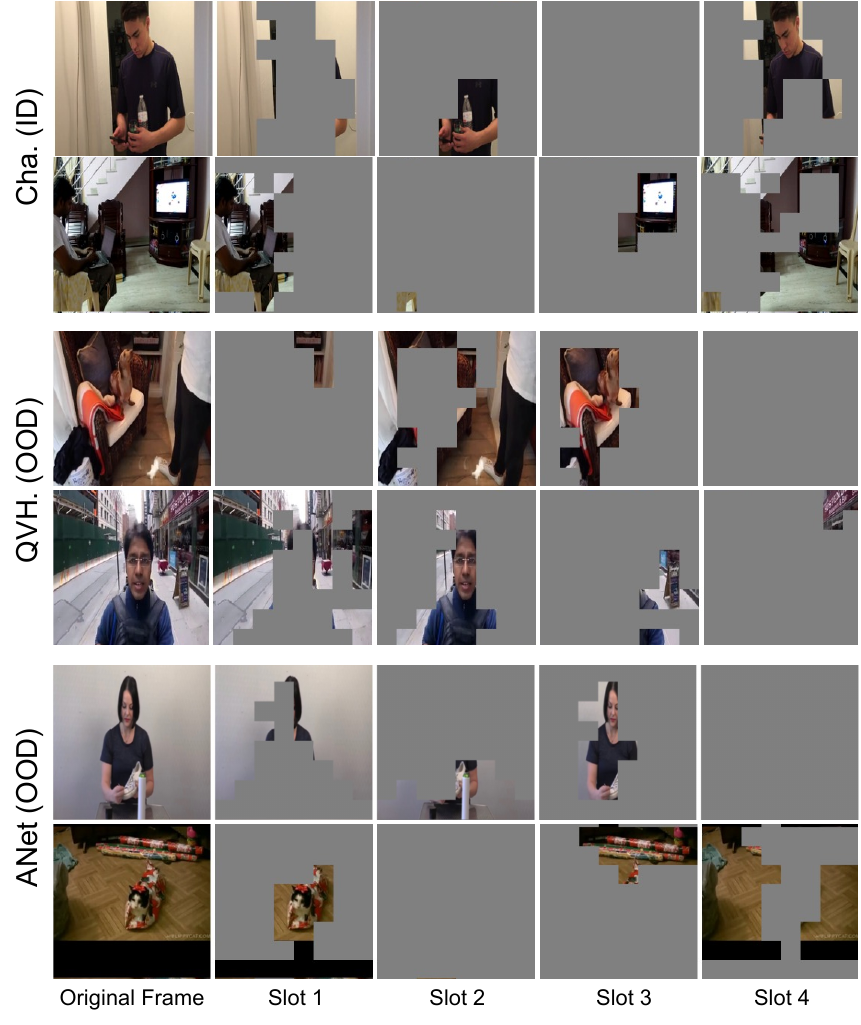}
\caption{\textbf{Slot attention visualization.} We visualize the slot assignments on samples from Cha.\ (ID), QVH.\ (OOD), and ANet (OOD) by masking each frame with its highest-attending slot. Each column corresponds to one of the four learned slots.}
\label{fig:slotviz}
\end{figure}

\noindent\textbf{What Do Slots Learn?}
We visualize the slot attention maps in \cref{fig:slotviz} by masking each frame region with its highest-attending slot. Across both ID and OOD samples, the slots decompose scenes into semantically coherent regions such as people, objects, and backgrounds, though the specific slot-to-entity mapping varies across frames. Importantly, this decomposition generalizes to unseen domains (QVH., ANet) without any domain-specific supervision, confirming that the Slot Adapter learns transferable entity-level representations rather than dataset-specific patterns.

\input{table/ablation}

\subsection{Ablation Study}
\label{sec:ablation}

We conduct extensive ablation studies to verify the effectiveness of each component in SlotVTG (\cref{tab:ablation_ours}). Unless otherwise stated, we use Qwen2.5-VL-3B~\cite{bai2025qwen25vl} as the backbone and train on Cha., reporting both ID (Cha.) and OOD (ANet) performance in R1@0.5 and R1@0.7.

\noindent\textbf{Effects of Slot Adapter.}
We compare our Slot Adapter against two baselines: LoRA-only fine-tuning and an adapter with standard self-attention instead of slot attention (\cref{tab:ablation_ours}a). While all three achieve comparable ID performance, the Slot Adapter yields the best OOD performance, confirming that the competitive slot decomposition mechanism is key to improving generalization.

\noindent\textbf{Effects of SA Loss.}
Removing the SA loss noticeably degrades out-of-distribution (OOD) performance (28.0 vs. 28.7 in R1@0.5), as shown in \cref{tab:ablation_ours}b. While increasing $\lambda$ to 0.2 improves in-distribution (ID) R1@0.5 performance to 64.3, it harms OOD performance, dropping it to 26.1. 
This trade-off suggests that enforcing an excessively strong objectness prior may lead to overfitting on source-domain patterns. 
Therefore, we set $\lambda$ = 0.1 as our default value.

\noindent\textbf{Effects of Slot Adapter Insertion Layers.}
Integrating the Slot Adapter into the early layers (1--7) yields the best out-of-distribution (OOD) performance, as shown in \cref{tab:ablation_ours}c. 
This aligns with the findings of~\cite{maptheflow}, which demonstrate that cross-frame interactions predominantly occur in the early decoder layers.
Conversely, applying the adapter to the middle (10--17) or later (20--36) layers degrades OOD performance, indicating that late-stage interventions likely introduce unnecessary noise.

\noindent\textbf{Effects of Number of Slots and Bottleneck Dimensions.}
We vary the number of slots $N_s$ and bottleneck dimension $d$ (\cref{tab:ablation_ours}d). Using $N_s{=}4$ and $d{=}512$ achieves the best OOD performance. A smaller dimension ($d{=}128$) slightly improves ID but degrades OOD, while increasing to $N_s{=}8$ slots hurts ID, likely because excessive slots dilute the decomposition.

\noindent\textbf{Effects of Token Reconstruction Design.}
We compare two strategies for reconstructing the original token sequence from the $N_s$ slots (\cref{tab:ablation_ours}e): (1) repeating each slot $N/N_s$ times followed by a linear projection, and (2) cross-attention where original tokens query the slots. Cross-attention achieves better OOD performance (29.3 vs.\ 28.2 R1@0.5), as it allows each token to selectively retrieve entity-aware information from its most relevant slot rather than receiving a uniform representation.

\noindent\textbf{Effects of SA Loss Placement.}
Applying the SA loss only at the last adapter layer (layer 7) outperforms applying it across all adapter layers (1--7) in OOD 
(\cref{tab:ablation_ours}f).
While full-layer constraints force premature alignment, applying them only to the last layer allows earlier layers to learn more flexible representations.

%% file: table/ablation.tex
\begin{table*}[t]
\centering

\caption{\textbf{Ablation study.} 
To validate the effect of each component in SlotVTG, we show the results on the Cha.$\rightarrow$ANet setting.
`Cha.' and `ANet' denote Charades-STA~\cite{gao2017tall} and ActivityNet Captions~\cite{krishna2017dense}.
We use Qwen2.5-VL~\cite{bai2025qwen25vl} 3B as a backbone MLLM.  
We report R1@0.5 and R1@0.7 scores on both ID and OOD settings.
The best numbers are \textbf{highlighted}.
}

\mpage{0.48}{
(a) Effects of Slot Adapter.
\\
\resizebox{\linewidth}{!}{
\begin{tabular}{l c cc c cc}
\toprule
\multirow{2}{*}{Adapter Type} && \multicolumn{2}{c}{Cha. (ID)} & 
& \multicolumn{2}{c}{ANet (OOD)} \\ 
\cline{3-4} \cline{6-7} 
&& R1@0.5 & R1@0.7 && R1@0.5 & R1@0.7 \\
\midrule
LoRA~\cite{hu2022lora} && 63.4 & 40.3 && 26.3 & 13.1 \\
Adapter w/ self attention && 63.5 & 40.9 && 26.5 & 13.7 \\ 
Slot Adapter && \textbf{64.0} & \textbf{41.2} && \textbf{28.7} & \textbf{14.4} \\
\bottomrule
\end{tabular}
}
}
\mpage{0.48}{
(b) Effects of SA loss.
\\
\resizebox{\linewidth}{!}{
\begin{tabular}{c c c cc c cc}
\toprule
\multirow{2}{*}{$\mathcal{L}_{SA}$} & \multirow{2}{*}{Loss scale $\lambda$} && \multicolumn{2}{c}{Cha. (ID)} & 
& \multicolumn{2}{c}{ANet (OOD)} \\ 
\cline{4-5} \cline{7-8} 
&&& R1@0.5 & R1@0.7 && R1@0.5 & R1@0.7 \\
\midrule
$\times$ & && 63.3 & 41.0 && 28.0 & 14.0 \\
$\checkmark$ & 0.1 && 64.0 & 41.2 && \textbf{28.7} & \textbf{14.4} \\
$\checkmark$ & 0.2 && \textbf{64.3} & \textbf{42.2} && 26.1 & 13.0 \\
\bottomrule
\end{tabular}
}
}
\mpage{0.43}{
(c) Effects of Slot Adapter insertion layers.
\\
\resizebox{\linewidth}{!}{
\begin{tabular}{l c cc c cc}
\toprule
\multirow{2}{*}{Layer index $l$} && \multicolumn{2}{c}{Cha. (ID)} & 
& \multicolumn{2}{c}{ANet (OOD)} \\ 
\cline{3-4} \cline{6-7} 
&& R1@0.5 & R1@0.7 && R1@0.5 & R1@0.7 \\
\midrule
1-7 && 63.3 & 41.0 && \textbf{28.7} & \textbf{14.4} \\
10-17 && 63.3 & 41.0 && 27.5 & 14.0 \\
20-36 && 63.3 & 39.5 && 28.4 & 14.0 \\
\bottomrule
\end{tabular}
}
}
\mpage{0.52}{
(d) Effects of number of slots and bottleneck dimensions.
\\
\resizebox{\linewidth}{!}{
\begin{tabular}{c c c cc c cc}
\toprule
\multirow{2}{*}{\# of slots $N_s$} & \multirow{2}{*}{Dimension $d$} && \multicolumn{2}{c}{Cha. (ID)} & 
& \multicolumn{2}{c}{ANet (OOD)} \\ 
\cline{4-5} \cline{7-8} 
&&& R1@0.5 & R1@0.7 && R1@0.5 & R1@0.7 \\
\midrule
4 & 128 && 64.1 & 41.7 && 28.5 & 14.1 \\
4 & 512 && 63.3 & 41.0 && \textbf{28.7} & \textbf{14.4} \\
8 & 512 && 63.7 & 39.8 && 28.6 & 14.5 \\
\bottomrule
\end{tabular}
}
}
\mpage{0.48}{
(e) Effects of token reconstruction design.
\\
\resizebox{\linewidth}{!}{
\begin{tabular}{l c cc c cc}
\toprule
\multirow{2}{*}{Method} && \multicolumn{2}{c}{Cha. (ID)} & 
& \multicolumn{2}{c}{ANet (OOD)} \\ 
\cline{3-4} \cline{6-7} 
&& R1@0.5 & R1@0.7 && R1@0.5 & R1@0.7 \\
\midrule
Repeat \& Proj. && \textbf{63.6} & 40.8 && 28.2 & 13.7 \\
Cross attention && 63.3 & \textbf{41.0} && \textbf{29.3} & \textbf{14.9} \\
\bottomrule
\end{tabular}
\label{tab:ablation_ours}
}
}
\mpage{0.48}{
(f) Effects of SA loss placement.
\\
\resizebox{\linewidth}{!}{
\begin{tabular}{l c cc c cc}
\toprule
\multirow{2}{*}{Layer index $l$} && \multicolumn{2}{c}{Cha. (ID)} & 
& \multicolumn{2}{c}{ANet (OOD)} \\ 
\cline{3-4} \cline{6-7} 
&& R1@0.5 & R1@0.7 && R1@0.5 & R1@0.7 \\
\midrule
1-7 (all adapter layers) && \textbf{64.0} & \textbf{41.5} && 28.5 & 14.3 \\
7 (last adapter layer) && \textbf{64.0} & 41.2 && \textbf{29.3} & \textbf{14.9} \\
\bottomrule
\end{tabular}
}
}
\label{tab:ablation}
\end{table*}

%% file: 5_conclusion.tex
\section{Conclusion}
\label{sec:conclusion}

We presented SlotVTG, a parameter-efficient framework that introduces object-centric decomposition into pre-trained MLLMs for generalizable Video Temporal Grounding. Our failure analysis reveals that na\"ively fine-tuned MLLMs exploit dataset-specific shortcuts rather than grounding in visual content. SlotVTG addresses this via a lightweight Slot Adapter that decomposes visual tokens into entity-level slots in the early decoder layers, guided by a Slot Alignment Loss that distills objectness priors. Extensive experiments demonstrate that SlotVTG consistently improves OOD generalization while maintaining competitive ID performance. 